\documentclass[final]{article}

\usepackage{microtype}
\usepackage{graphicx}
\usepackage{subcaption}
\usepackage{booktabs} 

\usepackage{hyperref}

\usepackage[accepted]{icml2025}

\usepackage{amsmath}
\usepackage{amssymb}
\usepackage{mathtools}
\usepackage{amsthm}

\usepackage[capitalize,noabbrev]{cleveref}

\theoremstyle{plain}
\newtheorem{theorem}{Theorem}[section]
\newtheorem{proposition}[theorem]{Proposition}

\theoremstyle{definition}

\theoremstyle{remark}

\usepackage{bm}
\usepackage{multirow}
\usepackage{tablefootnote}

\usepackage{xcolor}

\usepackage{setspace}

\newcommand{\etal}{~et~al.}

\usepackage{array}
\newcommand{\PreserveBackslash}[1]{\let\temp=\\#1\let\\=\temp}
\newcolumntype{C}[1]{>{\PreserveBackslash\centering}p{#1}}
\newcolumntype{R}[1]{>{\PreserveBackslash\raggedleft}p{#1}}
\newcolumntype{L}[1]{>{\PreserveBackslash\raggedright}p{#1}}

\usepackage{colortbl}
\definecolor{Gray}{gray}{0.935}
\newcolumntype{g}{>{\columncolor{Gray}}c}

\newcommand{\mynorm}[1]{\Vert#1\Vert_2}

\newcommand{\weight}{\bm{\pi}}

\begin{document}

\twocolumn[
\icmltitle{Bayesian Robust Aggregation for Federated Learning}



\icmlsetsymbol{equal}{*}

\begin{icmlauthorlist}
\icmlauthor{Aleksandr Karakulev}{equal,yyy}
\icmlauthor{Usama Zafar}{equal,yyy}
\icmlauthor{Salman Toor}{yyy,comp}
\icmlauthor{Prashant Singh}{yyy,sch}
\end{icmlauthorlist}

\icmlaffiliation{yyy}{Department of Information Technology, Uppsala University, Sweden}
\icmlaffiliation{sch}{Science for Life Laboratory, Sweden}
\icmlaffiliation{comp}{Scaleout Systems, Uppsala, Sweden}

\icmlcorrespondingauthor{Aleksandr Karakulev}{aleksandr.karakulev@it.uu.se}


\vskip 0.3in
]



\printAffiliationsAndNotice{\icmlEqualContribution} 

\begin{abstract}
Federated Learning enables collaborative training of machine learning models on decentralized data. This scheme, however, is vulnerable to adversarial attacks, when some of the clients submit corrupted model updates. In real-world scenarios, the total number of compromised clients is typically unknown, with the extent of attacks potentially varying over time. To address these challenges, we propose an adaptive approach for robust aggregation of model updates based on Bayesian inference. The mean update is defined by the maximum of the likelihood marginalized over probabilities of each client to be `honest'. As a result, the method shares the simplicity of the classical average estimators (e.g., sample mean or geometric median), being independent of the number of compromised clients. At the same time, it is as effective against attacks as methods specifically tailored to Federated Learning, such as Krum. We compare our
approach with other aggregation schemes in federated setting on three benchmark image classification data sets. The proposed method consistently achieves state-of-the-art performance across various attack types with static and varying number of malicious clients.
\end{abstract}
\section{Introduction}
\label{section:introduction}

Federated learning (FL) has emerged as an effective paradigm for privacy-preserving machine learning, enabling multiple clients, such as hospitals or banks, to collaboratively train a global model without sharing raw data \cite{mcmahan2017communication}. Instead of collecting sensitive data (e.g., medical records) centrally, clients share only iterative model updates with a central server, which aggregates these updates to refine the global model. This decentralized approach tackles key privacy issues and lowers the risks linked to data breaches, making FL highly attractive for both industry and academic applications ~\cite{teo2024federated, dayan2021federated, cheng2021secureboost, ye2020edgefed, jiang2020customized}. FL also aligns closely with the principle of data minimization, which is a core element in AI-related regulations such as GDPR and the AI Act \cite{TRUONG2021102402, woisetschläger2024federatedlearningprioritieseuropean}. 

\begin{figure}[t]
    \vskip 0.2in
    \begin{center}
        \includegraphics[width=0.9\columnwidth]{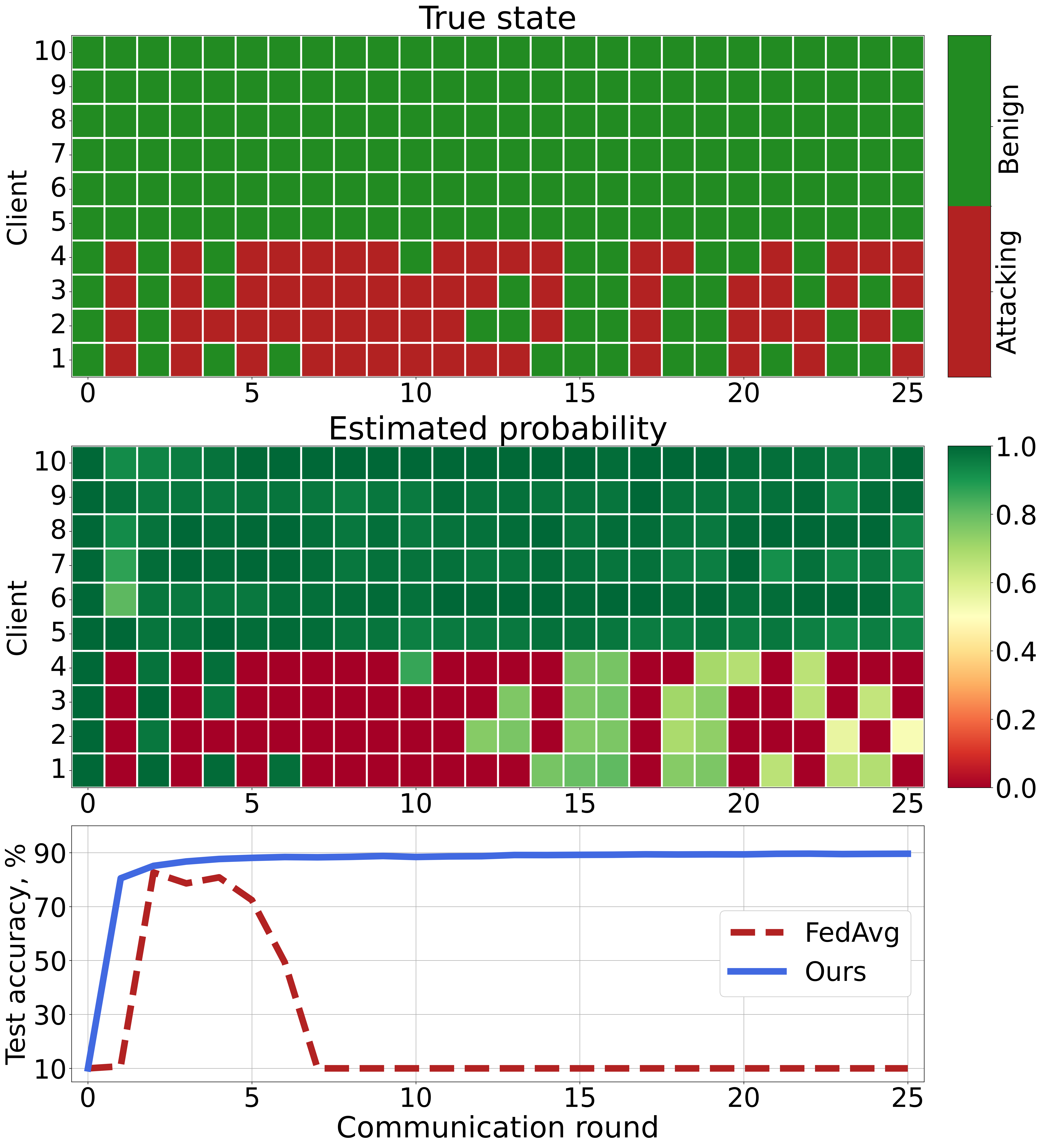}
        \caption{
            Federated Learning for classifying FMNIST data using 25 rounds and 10 clients. 
            Attacks are dynamic: in some communication rounds, malicious clients submit honest updates (\emph{upper heatmap}).
            Our aggregation method estimates probability for each client of being `benign' (\emph{lower heatmap}), which results in a global model with high test accuracy (\emph{bottom}).
        }
        \label{fig:illustration}
    \end{center}
    \vskip -0.2in
\end{figure}

Formally, in each communication round $t$, the central server collects model parameters $\bm{w}_k^{t}$ from $K$ clients, trained on their local datasets. The server then aggregates the collected updates using an aggregation mechanism $\mathcal{A}$, e.g., a weighted mean as in Federated Averaging (\textsc{FedAvg})~\cite{mcmahan2017communication}, to obtain the updated global model:
\begin{equation}
\bm{w}^{t+1} = \mathcal{A}(\bm{w}_1^{t}, \bm{w}_2^{t}, \ldots, \bm{w}_K^{t}).
\end{equation}
This iterative process continues until convergence.

However, the decentralized nature of FL introduces significant security challenges. In the presence of \emph{malicious clients} that intentionally disrupt the learning process, the integrity of the global model is at risk. Malicious clients may submit manipulative model updates, leading to a sub-optimal or even a harmful global model. See an example in \cref{fig:illustration} for training classifier on FMNIST dataset~\cite{xiao2017fmnist}, with some model updates being corrupted (by a sign-flip attack). Moreover, the number of compromised nodes changes during the 25 communication rounds. Therefore, a standard method like \textsc{FedAvg} fails. The proposed method, in contrast, is based on marginalizing over probabilities of each client being compromised at each communication round. This Bayesian approach makes the global training robust and adaptive, while allowing for a straightforward interpretation of the involved probabilistic variables, as can be seen from the heatmaps in \cref{fig:illustration}.

Several other adaptations to the federated averaging process have been proposed to enable robustness. Techniques such as \emph{Krum}~\cite{blanchard2017machine} were developed for aggregating models in the adversarial FL setting. Additionally, some classical estimators, such as \emph{Trimmed Mean}~\cite{yin2018byzantine} and \emph{Geometric Median}~\cite{pillutla2022Robust} have been explored in FL to mitigate the impact of malicious updates as well. These methods aim to identify and exclude outliers or adversarial contributions during aggregation. However, many existing approaches struggle to balance robustness with model performance,
or require an estimate of the number of compromised clients.

To address the challenges outlined above, we propose a principled and parameter-free robust aggregation method based on Bayesian inference \cite{karakulev2024adaptive}, that is general in application, simple in construction and applicable to both i.i.d and non-i.i.d FL settings. 
We demonstrate the method on a variety of attack types and proportion of compromised clients.

\section{Method}
In this section, we define the robust aggregation for Federated Learning based on Bayesian inference.
To this end, we formulate the averaging procedure as the maximum likelihood estimation in the view of contaminated observations.

\textbf{Robust aggregation}. 
One of the simplest aggregation schemes is the classical sample mean, wherein at round $t+1$, the weights $\bm{w}$ of the global model are updated as 
\begin{equation}
    \bm{w}^{t+1} = \left( \bm{w}^t_1 + \bm{w}^t_2 + \hdots + \bm{w}^t_K \right) / K.
    \label{eq:sample-mean}
\end{equation}
From the statistical view, it is equivalent to maximizing the Gaussian likelihood:
\begin{align}
    \bm{w}^{t+1} &= \arg\max_{\bm{w}} \prod_{k=1}^{K}\exp{\left( -\mynorm{\bm{w}_k^t - \bm{w}}^2 \right)}.
    \label{eq:likelihood}
\end{align}

However, malicious clients that submit incorrect model weights make this standard estimator impractical. To make the aggregation rule reliable under the presence of compromised clients, we consider a different -- robust -- formulation. Note that we address robust aggregation at each communication round $t+1$ as an independent task to keep the derivation general and thus omit the index $t$ below.

Consider $K$ vectors $\bm{w}_1, \bm{w}_2, ..., \bm{w}_K \in \mathbb{R}^d$ of which $M < K / 2$ are \emph{malicious}. We formulate the robust aggregation rule as computing the mean vector 
\begin{align}
    \overline{\bm{w}}_S := \frac{1}{\vert S \vert}\sum_{k \in S} \bm{w}_k
\end{align}
from a  subset of points defined by $S \subseteq \{1, 2, \dots, K\}$, such that $\vert S \vert = K - M$ and the sum of squared distances between that $\overline{\bm{w}}_S$ and vectors $\{\bm{w}_k\}_{k \in S}$ is minimized:
\begin{align}
    \min_{\substack{S \subseteq \{1,...,K\} \\ \vert S \vert = K-M}} \sum_{k \in S} \Vert \bm{w}_k - \overline{\bm{w}}_S \Vert_2^2.
    \label{eq:set-form}
\end{align}
Notice that \cref{eq:set-form} is equivalent to maximizing the Gaussian likelihood over a subset of some selected observations with respect to both the location parameter and the subset of points itself.

This formulation, however, leads to a combinatorially hard problem due to the need to check each possible subset of the given cardinality. Subsequently, we will show how \cref{eq:set-form} can be relaxed to Bayesian inference that brings two advantages: first, we get an efficient way to solve the problem and, second, we can optimize the number of malicious clients $M$, when it is unknown, and therefore make the aggregation adaptive.

But first, we motivate \cref{eq:set-form}. In fact, it yields a solution that satisfies the definition of an $(M, \kappa)-$robust aggregation rule — a criterion for a `good' averaging estimator that is commonly used in the Federated Learning literature \cite{allouah2023fixing,gorbunovvariance,karimireddybyzantine}.
\begin{proposition} 
For any vectors $\bm{w}_1, ..., \bm{w}_K \in \mathbb{R}^d$ and any set $B \subseteq \{1, ..., K\}$ of cardinality $K - M$, where $M < K/2$, denote  $\overline{\bm{w}}_B = \sum_{k \in B} w_k/\vert B \vert$. Then for solution $\overline{\bm{w}}_S$ of \cref{eq:set-form}, we have
\begin{align}
    \Vert \overline{\bm{w}}_S - \overline{\bm{w}}_B \Vert^2 \leq \dfrac{\kappa}{K-M} \sum_{k \in B} \Vert \bm{w}_k - \overline{\bm{w}}_B \Vert^2
    \label{eq:robust-agg}
\end{align}
where $\kappa = 4 \left( 1 + \dfrac{M}{K - 2M} \right)$.
\label{proposition}
\end{proposition}
One can compare this result with similar bounds from \cite{allouah2023fixing} for other widely used aggregation rules. The proof of the proposition is given in \cref{section:appendix-analysis}.

\textbf{Bayesian relaxation}. To make the problem in \cref{eq:set-form} tractable, we use Bayesian inference for `averaging' over the possible choice of subset $S$.
To this end, we express the problem in terms of indicator variables instead of using a set-based description:
\begin{align}
\min_{\substack{b_1, ..., b_K \in \{0, 1\} \\ \sum b_k = K-M}} \sum_{k=1}^K b_k \, \Vert \bm{w}_k - \overline{\bm{w}} \Vert_2^2, 
\label{eq:indicator-form}
\\
\overline{\bm{w}} = \dfrac{1}{\sum_k b_k} \sum_{k=1}^K b_k \bm{w}_k.\nonumber
\end{align}

But this problem is equivalent to
\begin{align}
    \max_{\overline{\bm{w}} \in \mathbb{R}^d}\max_{\substack{b_1, ..., b_K \in \{0, 1\} \\ \sum b_k = K-M}} \sum_{k=1}^K p(\bm{w}_k | \overline{\bm{w}})^{b_k},
    \label{eq:likelihood-form}
\end{align}
if the Gaussian likelihood $p(\bm{w}_k | \overline{\bm{w}}) \propto \exp(-\Vert \bm{w}_k - \overline{\bm{w}} \Vert^2)$ is used. 
The latter formulation gives a statistical view on the original problem defined in \cref{eq:set-form}. It allows us to treat the indicators $b_k$ as latent variables and perform \emph{marginalization} over them. That is, we relax an integer problem in \cref{eq:likelihood-form} using a convex combination
\begin{align}
    \max_{\overline{\bm{w}} \in \mathbb{R}^d} \mathbb{E}_{b_1, ..., b_k} \prod_{k=1}^K p(\bm{w}_k | \overline{\bm{w}})^{b_k},
    \label{eq:marginalized-form}
\end{align}
defined by the `prior' distribution 
\begin{align}
    p(b_1, ..., b_K) = \prod_{k=1}^K (1 - \varepsilon)^{b_k} \varepsilon^{1 - b_k}, \quad \varepsilon = M/K
\end{align}
that corresponds to only $K-M$ vectors being \emph{benign}.
Furthermore, the objective in \cref{eq:marginalized-form} can formally be optimized with respect to $\varepsilon$ as well, which makes the final algorithm adaptive. However, direct maximization of \cref{eq:marginalized-form} is still intractable. 

Subsequently we follow \cite{karakulev2024adaptive} in which the authors consider the general problem of robust likelihood maximization. They propose to maximize the marginal likelihood of the form \eqref{eq:marginalized-form} using variational posterior parameterized by $\weight = (\pi_1,\hdots, \pi_K)$. Consequently, $0 \leq \pi_k \leq 1$ estimates the \emph{posterior} probability that $b_k = 1$. This approximation leads to the following \emph{evidence lower bound} (ELBO):
\begin{flalign}
    &\max_{\bm{w},\, \weight} \; \sum\limits_{k=1}^{K} \pi_k \ln p(\bm{w}_k | \overline{\bm{w}}) - \mathrm{KL}(\weight, \varepsilon), 
    \quad \text{where}
    \label{eq:objective}
    \\ 
    &\mathrm{KL}(\weight, \varepsilon) = \sum\limits_{k=1}^K \left[\pi_k \ln\frac{\pi_k}{1 - \varepsilon} + (1 - \pi_k) \ln \frac{1 - \pi_k}{\varepsilon}
    \right]. 
    \nonumber    
\end{flalign}
The first term is the log-likelihood averaged over posterior probabilities $\pi_k$. The second term $\mathrm{KL}(\weight, \varepsilon)$ corresponds to the KL-divergence between the posterior defined by $\pi_k$ and the prior defined by $\varepsilon$.
Note that the unknown hyperparameter $\varepsilon$ is thus optimized explicitly from the KL-term:
\begin{align}
    \varepsilon = 1 - \sum_{k=1}^K \pi_k / K.
\end{align}
Such hyperparameter-free inference (known as `empirical Bayes' \cite{murphy2012machine}) can be useful in dynamic settings such as Federated Learning, when the number of compromised clients $M = \varepsilon K$ may not be constant.

We note that in \cite{karakulev2024adaptive} the authors motivate maximization of some likelihood marginalized over indicator variables by considering the data from a mixture of two distributions, $(1-\varepsilon)P + \varepsilon Q$, which is a standard contamination model from robust statistics \cite{huber1996robust}. In such a consideration, it is implied that `benign' points come from the common `true' component of the mixture $P$, and then, the likelihood corresponding to distribution $P$  is used in \eqref{eq:objective}. However, in Federated Learning we can only assume independence of `benign' model updates $\bm{w}_k$, but they may not be identically distributed due to a heterogeneous distribution of the data. 
Thus, in contrast to the i.i.d. setting considered in \cite{karakulev2024adaptive}, we motivate the usage of the simple Gaussian likelihood in the ELBO \eqref{eq:objective} by showing that this likelihood yields a robust aggregation rule in \cref{eq:set-form}.

Also, we remark that formally the marginalized likelihood in \eqref{eq:marginalized-form} is considered as a tractable relaxation of the original combinatorial problem. However, it is often observed in practice that marginalizing over unknown variables, such as indicators $b_k$, leads to more reasonable, less extreme solutions (Bayesian approach)  compared to finding the most plausible values of these variables and using them for the final estimate (frequentist approach)  \cite{mackay2003information}. Our empirical results indicate that the optimum of the marginal likelihood \eqref{eq:marginalized-form} proves effective, and notably, it also adapts to the number of malicious clients (see \cref{fig:illustration}).

\textbf{Numerical optimization}. Within the derived objective \eqref{eq:objective} for robust aggregation, we use the Gaussian likelihood according to \eqref{eq:set-form}. In general, this likelihood allows any (positive) variance, since the standard sample mean is invariant to the scale of the data points. In contrast, the evidence lower bound depends on the scale parameter. Namely, with different model parameterization, the term containing the residuals $\mynorm{\bm{w}_k - \overline{\bm{w}}}^2$ can grow arbitrarily large, while the term $\mathrm{KL}(\weight, \varepsilon)$ 
does not depend on the weight vectors $\bm{w}_k$.
To allow for consistency of the objective function, we normalize the Euclidean distances by the variance $\sigma^2$.
In other words, we optimize the likelihood term of the ELBO with respect to two parameters: location and scale, $\bm{\theta} = (\overline{\bm{w}}, \sigma^2)$. Consequently, the first term in the ELBO becomes:
\begin{align}
    \sum\limits_{k=1}^{K} \pi_k \ln p(\bm{w}_k | \bm{\theta})
    = -\sum\limits_{k=1}^K \frac{\pi_k}{2} \left[ \dfrac{\mynorm{\bm{w}_k - \overline{\bm{w}}}^2}{\sigma^2} + \ln({2\pi \sigma^2}) \right],
    \label{eq:first-term}
\end{align}
while the KL-term $\mathrm{KL}(\weight, \varepsilon)$ remains untouched.
Given fixed weights $\pi_k$, the location and scale parameters can be obtained in the closed form from \cref{eq:first-term}:
\begin{align}
    \overline{\bm{w}} = \dfrac{\sum_{k=1}^K \pi_k \,\bm{w}_k}{\sum_{k=1}^K \pi_k}, \quad 
    \sigma^2 = \dfrac{\sum_{k=1}^K \pi_k \mynorm{\bm{w}_k - \overline{\bm{w}}}^2}{\sum_{k=1}^K \pi_k}.
    \label{eq:robust-loc-scale}
\end{align}
Subsequently, posterior probabilities $\pi_k$ of `benign' vs. `malicious' labels can be updated for the given mean and variance, using iterations~\cite{karakulev2024adaptive}: for $k=1,\hdots,K$,
\begin{align}
    \pi_k^{\text{new}} = \left( 1 + \dfrac{K - \sum_k \pi_k^\text{old}}{\sum_k \pi_k^\text{old}}\dfrac{1}{p(\bm{w}_k | \overline{\bm{w}}, \sigma^2)} \right)^{-1}.
\label{eq:fixed-point}
\end{align}
Taken together, step-wise optimization of $(\overline{\bm{w}}, \,\sigma^2)$ and $\weight$ results into the robust average of the model updates for Federated Learning in the setting with compromised clients. The procedure is listed as \cref{alg:aggregation}.
\begin{algorithm}[t]
    \setstretch{1.3}
    \caption{\textsc{Bayesian Robust Aggregation}}
    \label{alg:aggregation}
 \begin{algorithmic}[1]
    \STATE {\bfseries Input:} local models $\bm{w}^t_1, \hdots, \bm{w}^t_K$ from round $t$
    \STATE $\pi_k = 1$ \textbf{for} $k=1,\dots,K$
    \STATE $\overline{\bm{w}} = \sum_{k=1}^K \bm{w}^t_k \,/\, K$
    \STATE $\sigma^2 = \sum_{k=1}^K \mynorm{\bm{w}^t_k - \overline{\bm{w}}}^2 \,/\, K $
    \REPEAT
        \STATE update $\weight \leftarrow$ with \cref{eq:fixed-point} using $\overline{\bm{w}}$ and $\sigma^2$
        \STATE update $\overline{\bm{w}}  \leftarrow \sum_{k=1}^K \pi_k \bm{w}^t_k \,/\, \sum_{k=1}^K \pi_k$
        \STATE update $\sigma^2 \leftarrow \sum_{k=1}^K \pi_k \mynorm{\bm{w}^t_k - \overline{\bm{w}}}^2 \,/\, \sum_{k=1}^K \pi_k$
    \UNTIL convergence
 \end{algorithmic}
 \end{algorithm}

\textbf{Computational complexity}.
Our final aggregation rule amounts to optimizing the ELBO in \eqref{eq:objective}, which corresponds to the evidence lower bound. Since \cref{alg:aggregation} is an EM-style (coordinate ascent) procedure, we can guarantee that it converges to a local maximum of the ELBO and indirectly maximizes the marginal likelihood in \eqref{eq:marginalized-form} \cite{blei2017variational}. While the number of iterations depends on the data distribution, we emphasize that each iteration is computationally efficient: the robust mean step for $\overline{\bm{w}}$ is simply a weighted mean (with the same cost as a sample mean over $K$ points), and updating the weights $\pi_k$ involves  independent scalar updates, as shown in \cref{eq:fixed-point}.
Overall, the complexity is $O(TK)$, where $T$ is the number of EM iterations.  
Importantly, since $T$ is fixed and there is no quadratic scaling in $K$, our method remains comparable in efficiency to, e.g., coordiante-wise median which has complexity $O(K)$. 
\section{Experimental Settings}
\label{section:experimet-settings}
We evaluate our approach against state-of-the-art robust aggregation algorithms in classification tasks with varying configurations. To ensure fair comparisons, we implement existing attacks and defenses following their original designs. The simulation framework is implemented using PyTorch~\cite{paszke2019pytorch}, and is easily extensible for further research\footnote{GitHub link: \href{https://github.com/SciML-FL/bra-fl}{\texttt{https://github.com/SciML-FL/bra-fl}}}. We describe the setup used to conduct the experiments in the following text. The results are presented in \cref{section:experiment-results}.

\subsection{Datasets and Model}
We focus on the image classification task using three benchmark datasets: \emph{MNIST}~\cite{deng2012mnist}, \emph{Fashion-MNIST (FMNIST)}~\cite{xiao2017fmnist}, and \emph{CIFAR-10}~\cite{krizhevsky2009learning}. The datasets represent a range of complexities, from simple digit recognition (MNIST) to more challenging tasks like object classification (CIFAR-10).

We use the \emph{LeNet-5}~\cite{lecun1998gradient} architecture for MNIST and Fashion-MNIST datasets, and \emph{ResNet-18}~\cite{he2015deep} for CIFAR-10 dataset, selecting these models based on their suitability for each dataset’s complexity. Further details about the datasets and model architectures are provided in \cref{section:appendix-datasets-models}.

Furthermore, the data is distributed among clients following a Dirichlet distribution, with $\alpha=1.0$ for i.i.d. and $\alpha=0.5$ for non-i.i.d. settings, similar to existing works by Zhang\etal~\yrcite{zhang2023flip} and Bagdasaryan\etal~\yrcite{bagdasaryanHowBackdoorFederated2019}.

\begin{table}[h]
    \vskip 0.1in
    \caption{Summary of malicious client configurations. The total number of clients ($N$) is set to 20 for each configuration.}
    \label{tab:configs-t1}
    \vskip 0.1in
    \begin{center}
        \begin{small}
            \begin{tabular}{cR{1cm}R{1cm}C{1.5cm}}\toprule[1.125pt]
                Config \#               & $\varepsilon$         & $\varepsilon N$          & Dynamic      \\ \midrule[1.125pt]
                 1                      &   $0.2$               & $4$                      & $\times$              \\ 
                 2                      &   $0.4$               & $8$                      & $\times$              \\ 
                 3                      &   $\leq 0.45$         & $\leq 9$                 & \checkmark            \\ \bottomrule[1.125pt]
            \end{tabular}
        \end{small}
    \end{center}
    \vskip -0.1in
\end{table}

\subsection{Federated Setup}
For all experiments, we set the total number of clients $N$ to $20$. Each client trains the local models for $10$ epochs using the Stochastic Gradient Descent (SGD) optimizer with Nesterov momentum and L2 regularization. For the local training, we use a fixed learning rate of $0.01$, weight decay of $10^{-4}$, and set the momentum coefficient to $0.9$. These hyperparameters were chosen based on their widespread use in federated learning literature and their effectiveness in preliminary experiments.

Training lasts $100$ communication rounds for the MNIST and Fashion-MNIST datasets, and $200$ communication rounds for the CIFAR-10 dataset. The number of communication rounds is chosen based on the complexity of each dataset, with fewer rounds for simpler datasets (MNIST and FMNIST) and more rounds for CIFAR-10. We set the total number of clients $N = 20$ in order to balance computational efficiency and realistic federated learning scenarios.  Extended details about the training hyperparameters and configuration are provided in \cref{section:appendix-federated-setup}.

\subsection{Attack Setup}

We evaluate the robustness of our proposed solution, beginning with a small number of malicious clients and progressively scaling up to the theoretical maximum of 45\%. We assume that an adversary can compromise a fraction $\varepsilon$ of the total clients. To demonstrate our method's robustness, we test with $\varepsilon = 0.20$, $\varepsilon = 0.40$, and $\varepsilon = 0.45$, representing moderate to extreme levels of adversarial influence. These values were chosen to cover a wide range of scenarios, from a small but significant fraction of compromised clients ($\varepsilon = 0.20$) to a near-majority adversarial influence ($\varepsilon = 0.45$). We note that recent studies emphasize that a high number of malicious clients may be unrealistic in real-world FL settings \cite{9833647}. 

We compare our approach against the state-of-the-art using both untargeted and targeted attacks. The configurations for malicious clients are summarized under \cref{tab:configs-t1} which includes fraction of compromised clients and whether adversaries dynamically alternate between benign and malicious behavior. Dynamic adversarial behavior is included to simulate real-world scenarios where adversaries intermittently attack to evade detection.

We implement the following attacks to evaluate our method's effectiveness:
\begin{itemize}
    \item \textbf{Random Update attack}: The adversary submits random noise sampled from a Gaussian distribution as its local model updates. The aim is to disrupt the global model convergence by introducing arbitrary updates.

    \item \textbf{Sign Flipping (SignFlip) attack}: The adversary flips the signs of the model gradient~\cite{karimireddy2020learning}, effectively performing gradient ascent instead of gradient descent. This maximizes the loss and prevents the model from converging.

    \item \textbf{Label Flipping (LabelFlip) attack}: The adversary flips the labels~\cite{fang_local_2020} of each training sample on all malicious clients. Specifically, we rotate all labels, i.e., set a label $y$ to be $y+1\mod Y$ for each sample where $Y$ is the total number of classes.

    \item \textbf{Backdoor attack}: The local datasets of compromised clients are altered by adding a trigger (e.g., a small pixel pattern) to the samples of a specific target class and modifying their labels. We always add the trigger to samples of class $0$ and change their labels to class $8$. This attack aims to create a backdoor in the global model, causing it to misclassify triggered samples.
\end{itemize}
These attacks were selected to cover a range of adversarial strategies, from simple noise injection to sophisticated data poisoning, ensuring a comprehensive evaluation of our method's robustness. Further details concerning the attack configurations can be found in \cref{section:appendix-atttack-details}.

\subsection{Compared Defense Baselines}
We compare our robust aggregation scheme against five different federated learning baselines, i.e., 
\begin{itemize}
    \item \textbf{Federated Average (\textsc{FedAvg})}~\cite{mcmahan2017communication}: Aggregates updates by taking their weighted mean. The model's performance under a benign setting using \textsc{FedAvg} serves as the baseline for all methods.
    \item \textbf{Median}~\cite{yin2018byzantine}: Computes the coordinate-wise median of updates to reduce the impact of outliers. Median is naturally robust against outliers, making it useful for estimating a robust aggregate.
    \item \textbf{Geometric Median (\textsc{GeoMed})}~\cite{pillutla2022Robust}: Finds the geometric median of updates, which is robust to adversarial contributions. Unlike the coordinate-wise median, \textsc{GeoMed} considers a holistic view of the update, treating all components together rather than individually. 
    \item \textbf{Trimmed Mean (\textsc{TrimAvg})}~\cite{yin2018byzantine}: Excludes a fraction of extreme updates before averaging. The fraction of updates to discard on each extreme is controlled by a hyperparameter $\beta$, which varies depending on the experiment.
    \item \textbf{Multi-Krum (\textsc{Krum})}~\cite{blanchard2017machine}: Selects a subset of updates that are closest to each other, excluding potential outliers. \textsc{Krum} retains only $L$ updates, with $L$ being a user-defined hyperparameter.
\end{itemize}
These baselines were selected to represent a range of robust aggregation strategies, from simple averaging to more sophisticated Byzantine-resilient methods. 

\subsection{Evaluation Metrics}
Similar to previous works \cite{zhang2023a3fl, fang2023vulnerability, zhang2023flip}, we use test accuracy (ACC) and attack success rate (ASR) to evaluate the effectiveness of adversarial attacks. These metrics were chosen to comprehensively assess both the utility and robustness of the global model.

For untargeted attacks aiming to degrade model performance, we assess ACC. A significant decrease in accuracy indicates successful disruption of the model’s performance.

In contrast, the Backdoor attack is evaluated using ASR, which measures the proportion of triggered samples that are misclassified as the target class (class $8$). A higher ASR reflects greater success in manipulating the model’s predictions.

\begin{table}[t]
    \vskip 0.1in
    \caption{Results showing ACC and ASR under Sign Flipping and Backdoor Attacks with $20\%$ malicious clients. \\ $^\ast$We use $\beta = 0.2$ for \textsc{TrimAvg} and $L = 16$ for \textsc{Krum}.}
    \label{tab:results-template1-signback-config1}
    \vskip 0.1in
    \begin{center}
        \begin{small}
        \resizebox{0.95\columnwidth}{!}{%
            \begin{tabular}{llgcggcc}\toprule[1.5pt]
                & \multirow{4}{*}{\textbf{Baseline}} & \multicolumn{2}{c}{\textbf{SignFlip}}  & \multicolumn{4}{c}{\textbf{Backdoor}} \\
                \cmidrule(lr){3-4} \cmidrule(lr){5-8}
                 &  &  \multicolumn{1}{c}{$\alpha = 0.5$}  & $\alpha = 1.0$ & \multicolumn{2}{c}{\textbf{$\alpha = 0.5$}}  & \multicolumn{2}{c}{\textbf{$\alpha = 1.0$}} \\
                \cmidrule(lr){3-3} \cmidrule(lr){4-4} \cmidrule(lr){5-6} \cmidrule(lr){7-8}
                 ~ & ~ & ACC       & ACC & ACC & ASR       & ACC & ASR \\ \midrule[1.125pt]
                \multirow{7}{*}{\rotatebox{90}{\textbf{MNIST}}} 
                & \textsc{FedAvg} & 0.10 & 0.10 & 0.99 & 0.98 & 0.99 & 0.98 \\
                ~ & \textsc{Median} & 0.99 & 0.99 & 0.99 & 0.01 & 0.99 & 0.00 \\
                ~ & \textsc{TrimAvg}* & 0.99 & 0.99 & 0.99 & 0.00 & 0.99 & 0.00 \\
                ~ & \textsc{GeoMed} & 0.99 & 0.99 & 0.99 & 0.00 & 0.99 & 0.00 \\
                ~ & \textsc{Krum}* & 0.99 & 0.99 & 0.99 & 0.00 & 0.99 & 0.00 \\ \cmidrule(lr){2-8}
                ~ & \textbf{Ours} & 0.99 & 0.99 & 0.99 & 0.00 & 0.99 & 0.00 \\ \midrule[1.125pt]

                \multirow{7}{*}{\rotatebox{90}{\textbf{FMNIST}}} 
                & \textsc{FedAvg} & 0.10 & 0.10 & 0.89 & 0.83 & 0.89 & 0.83 \\
                ~ & \textsc{Median} & 0.88 & 0.88 & 0.88 & 0.12 & 0.88 & 0.05 \\
                ~ & \textsc{TrimAvg}* & 0.88 & 0.88 & 0.88 & 0.35 & 0.89 & 0.43 \\
                ~ & \textsc{GeoMed} & 0.88 & 0.88 & 0.89 & 0.11 & 0.89 & 0.10 \\
                ~ & \textsc{Krum}* & 0.89 & 0.89 & 0.89 & 0.01 & 0.89 & 0.01 \\ \cmidrule(lr){2-8}
                ~ & \textbf{Ours} & 0.89 & 0.89 & 0.89 & 0.01 & 0.89 & 0.01 \\ \midrule[1.125pt]
                 
                \multirow{7}{*}{\rotatebox{90}{\textbf{CIFAR-10}}}
                & \textsc{FedAvg} & 0.10 & 0.10 & 0.92 & 0.99 & 0.92 & 0.99 \\
                ~ & \textsc{Median} & 0.85 & 0.85 & 0.91 & 0.47 & 0.91 & 0.40 \\
                ~ & \textsc{TrimAvg}* & 0.84 & 0.84 & 0.91 & 0.98 & 0.91 & 0.98 \\
                ~ & \textsc{GeoMed} & 0.86 & 0.86 & 0.91 & 0.15 & 0.91 & 0.15 \\
                ~ & \textsc{Krum}* & 0.90 & 0.90 & 0.91 & 0.08 & 0.91 & 0.08 \\ \cmidrule(lr){2-8}
                ~ & \textbf{Ours} & 0.89 & 0.90 & 0.91 & 0.08 & 0.91 & 0.08 \\
                 \bottomrule[1.5pt]
            \end{tabular}%
        }
        \end{small}
    \end{center}
    \vskip -0.1in
\end{table}

\begin{table}[t]
    \vskip 0.1in
    \caption{Results showing ACC and ASR under Sign Flipping and Backdoor Attacks with $40\%$ malicious clients. \\$^\ast$We use $\beta = 0.4$ for \textsc{TrimAvg} and $L = 12$ for \textsc{Krum}.}
    \label{tab:results-template1-signback-config2}
    \vskip 0.1in
    \begin{center}
        \begin{small}
        \resizebox{0.95\columnwidth}{!}{%
            \begin{tabular}{llgcggcc}\toprule[1.5pt]
                & \multirow{4}{*}{\textbf{Baseline}} & \multicolumn{2}{c}{\textbf{SignFlip}}  & \multicolumn{4}{c}{\textbf{Backdoor}} \\
                \cmidrule(lr){3-4} \cmidrule(lr){5-8}
                 &  &  \multicolumn{1}{c}{$\alpha = 0.5$}  & $\alpha = 1.0$ & \multicolumn{2}{c}{\textbf{$\alpha = 0.5$}}  & \multicolumn{2}{c}{\textbf{$\alpha = 1.0$}} \\
                \cmidrule(lr){3-3} \cmidrule(lr){4-4} \cmidrule(lr){5-6} \cmidrule(lr){7-8}
                 ~ & ~ & ACC       & ACC & ACC & ASR       & ACC & ASR \\ \midrule[1.125pt]
                \multirow{7}{*}{\rotatebox{90}{\textbf{MNIST}}} 
                & \textsc{FedAvg} & 0.10 & 0.10 & 0.99 & 1.00 & 0.99 & 1.00 \\
                ~ & \textsc{Median} & 0.97 & 0.97 & 0.41 & 0.03 & 0.10 & 0.00 \\
                ~ & \textsc{TrimAvg}* & 0.97 & 0.97 & 0.99 & 0.57 & 0.98 & 0.17 \\
                ~ & \textsc{GeoMed} & 0.97 & 0.97 & 0.99 & 0.00 & 0.99 & 0.00 \\
                ~ & \textsc{Krum}* & 0.99 & 0.99 & 0.99 & 0.00 & 0.99 & 0.00 \\ \cmidrule(lr){2-8}
                ~ & \textbf{Ours} & 0.99 & 0.99 & 0.99 & 0.00 & 0.99 & 0.00 \\ \midrule[1.125pt]

                \multirow{7}{*}{\rotatebox{90}{\textbf{FMNIST}}}
                & \textsc{FedAvg} & 0.10 & 0.10 & 0.90 & 0.98 & 0.90 & 0.97 \\
                ~ & \textsc{Median} & 0.84 & 0.84 & 0.89 & 0.97 & 0.90 & 0.96 \\
                ~ & \textsc{TrimAvg}* & 0.83 & 0.84 & 0.90 & 0.97 & 0.90 & 0.96 \\
                ~ & \textsc{GeoMed} & 0.85 & 0.84 & 0.89 & 0.95 & 0.89 & 0.95 \\
                ~ & \textsc{Krum}* & 0.89 & 0.89 & 0.89 & 0.01 & 0.89 & 0.01 \\ \cmidrule(lr){2-8}
                ~ & \textbf{Ours} & 0.89 & 0.89 & 0.89 & 0.01 & 0.89 & 0.01 \\ \midrule[1.125pt]
                 
                \multirow{7}{*}{\rotatebox{90}{\textbf{CIFAR-10}}}
                & \textsc{FedAvg} & 0.10 & 0.10 & 0.92 & 1.00 & 0.91 & 1.00 \\
                ~ & \textsc{Median} & 0.64 & 0.65 & 0.91 & 1.00 & 0.91 & 1.00 \\
                ~ & \textsc{TrimAvg}* & 0.64 & 0.63 & 0.91 & 1.00 & 0.91 & 1.00 \\
                ~ & \textsc{GeoMed} & 0.78 & 0.74 & 0.92 & 0.96 & 0.91 & 0.96 \\
                ~ & \textsc{Krum}* & 0.88 & 0.88 & 0.91 & 0.08 & 0.91 & 0.08 \\ \cmidrule(lr){2-8}
                ~ & \textbf{Ours} & 0.88 & 0.88 & 0.91 & 0.08 & 0.91 & 0.08 \\
                 \bottomrule[1.5pt]
            \end{tabular}%
        }
        \end{small}
    \end{center}
    \vskip -0.1in
\end{table}

\begin{table}[t]
    \vskip 0.1in
    \caption{Results showing ACC and ASR under Sign Flipping and Backdoor Attacks with malicious clients $\leq 45\%$. \\$^\ast$We use $\beta = 0.45$ for \textsc{TrimAvg} and $L = 11$ for \textsc{Krum}.}
    \label{tab:results-template1-signback-config3}
    \vskip 0.1in
    \begin{center}
        \begin{small}
        \resizebox{0.95\columnwidth}{!}{%
            \begin{tabular}{llgcggcc}\toprule[1.5pt]
                & \multirow{4}{*}{\textbf{Baseline}} & \multicolumn{2}{c}{\textbf{SignFlip}}  & \multicolumn{4}{c}{\textbf{Backdoor}} \\
                \cmidrule(lr){3-4} \cmidrule(lr){5-8}
                 &  &  \multicolumn{1}{c}{$\alpha = 0.5$}  & $\alpha = 1.0$ & \multicolumn{2}{c}{\textbf{$\alpha = 0.5$}}  & \multicolumn{2}{c}{\textbf{$\alpha = 1.0$}} \\
                \cmidrule(lr){3-3} \cmidrule(lr){4-4} \cmidrule(lr){5-6} \cmidrule(lr){7-8}
                 ~ & ~ & ACC       & ACC & ACC & ASR       & ACC & ASR \\ \midrule[1.125pt]
                \multirow{7}{*}{\rotatebox{90}{\textbf{MNIST}}} 
                & \textsc{FedAvg} & 0.10 & 0.10 & 0.99 & 1.00 & 0.99 & 1.00 \\
                ~ & \textsc{Median} & 0.99 & 0.99 & 0.99 & 0.00 & 0.10 & 0.00 \\
                ~ & \textsc{TrimAvg}* & 0.99 & 0.99 & 0.99 & 0.00 & 0.99 & 0.00 \\
                ~ & \textsc{GeoMed} & 0.99 & 0.99 & 0.99 & 0.00 & 0.99 & 0.00 \\
                ~ & \textsc{Krum}* & 0.99 & 0.99 & 0.99 & 0.00 & 0.99 & 0.00 \\ \cmidrule(lr){2-8}
                ~ & \textbf{Ours} & 0.99 & 0.99 & 0.99 & 0.00 & 0.99 & 0.00 \\ \midrule[1.125pt]

                \multirow{7}{*}{\rotatebox{90}{\textbf{FMNIST}}}
                & \textsc{FedAvg} & 0.10 & 0.10 & 0.89 & 0.93 & 0.89 & 0.95 \\
                ~ & \textsc{Median} & 0.88 & 0.88 & 0.89 & 0.46 & 0.89 & 0.60 \\
                ~ & \textsc{TrimAvg}* & 0.88 & 0.88 & 0.89 & 0.59 & 0.89 & 0.49 \\
                ~ & \textsc{GeoMed} & 0.88 & 0.88 & 0.89 & 0.25 & 0.89 & 0.13 \\
                ~ & \textsc{Krum}* & 0.89 & 0.89 & 0.89 & 0.01 & 0.89 & 0.01 \\ \cmidrule(lr){2-8}
                ~ & \textbf{Ours} & 0.89 & 0.89 & 0.89 & 0.01 & 0.89 & 0.01 \\ \midrule[1.125pt]
                 
                \multirow{7}{*}{\rotatebox{90}{\textbf{CIFAR-10}}}
                & \textsc{FedAvg} & 0.13 & 0.13 & 0.91 & 1.00 & 0.92 & 1.00 \\
                ~ & \textsc{Median} & 0.88 & 0.88 & 0.91 & 0.97 & 0.91 & 0.97 \\
                ~ & \textsc{TrimAvg}* & 0.88 & 0.88 & 0.91 & 0.97 & 0.91 & 0.97 \\
                ~ & \textsc{GeoMed} & 0.88 & 0.88 & 0.91 & 0.30 & 0.91 & 0.30 \\
                ~ & \textsc{Krum}* & 0.89 & 0.89 & 0.91 & 0.08 & 0.91 & 0.08 \\ \cmidrule(lr){2-8}
                ~ & \textbf{Ours} & 0.89 & 0.90 & 0.91 & 0.08 & 0.91 & 0.08 \\
                 \bottomrule[1.5pt]
            \end{tabular}%
        }
        \end{small}
    \end{center}
    \vskip -0.1in
\end{table}

\section{Results}
\label{section:experiment-results}
We evaluate our proposed Bayesian Robust Aggregation scheme and compare it to baseline defense mechanisms outlined in \cref{section:experimet-settings}. We focus on two prominent attack types: Sign Flipping and Backdoor attacks, while results for Label-Flipping and Random Update attacks are deferred to \cref{section:appendix-additional_results}, as all robust aggregation methods, including ours, perform comparably well under the latter scenarios.

\begin{figure*}[t]
    \vskip 0.1in
    \begin{center}
        \begin{subfigure}[b]{0.45\linewidth}
            \centering
            \includegraphics[width=0.86\linewidth]{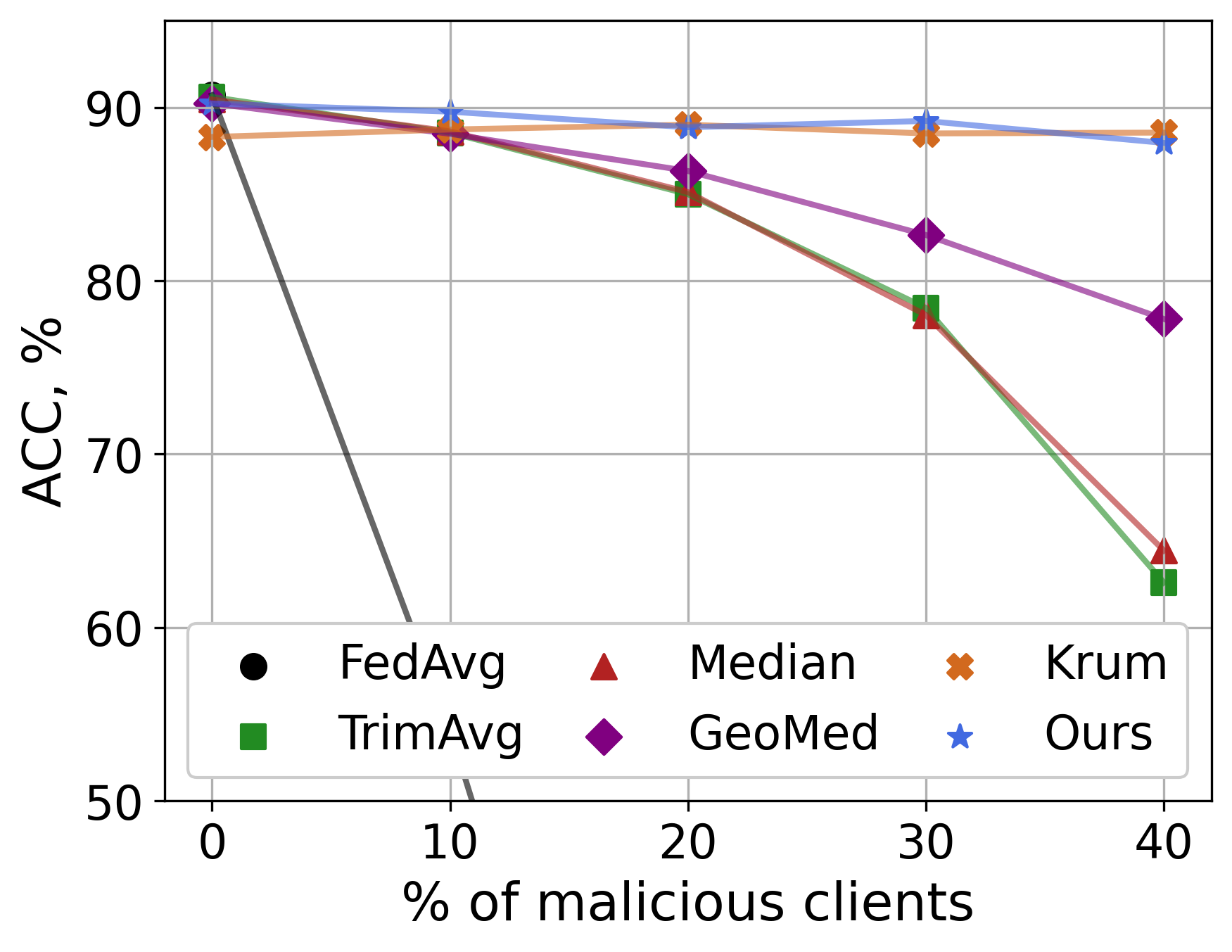}
            \caption{}
        \end{subfigure}%
        \begin{subfigure}[b]{0.45\linewidth}
            \centering
            \includegraphics[width=0.89\linewidth]{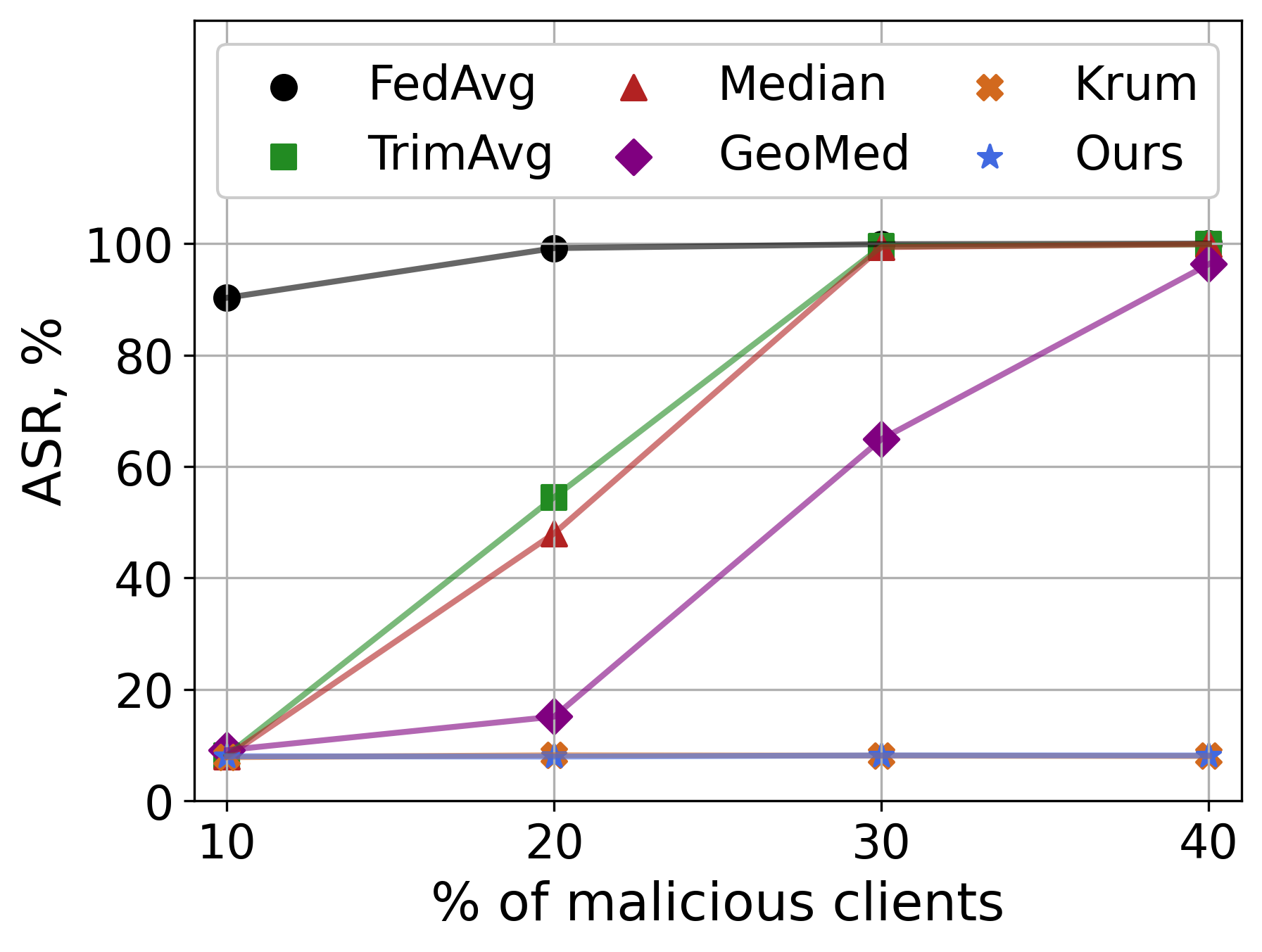}
            \caption{}
        \end{subfigure}\vfill
        \caption{Ablation study. (a) Test accuracy (ACC) under Sign Flip with fraction of malicious clients varying from 0 to 40\% and (b) attack success rate (ASR) under Backdoor attack with fraction of malicious clients varying from 10\% to 40\%.}
        \label{fig:ablation-sign_back}
    \end{center}
    \vskip -0.1in
\end{figure*}

\subsection{Evaluating Configurations 1--3}

\cref{tab:results-template1-signback-config1} summarizes the results for configuration 1, where 20\% of the clients are malicious. For the \emph{SignFlip attack}, which is an untargeted attack, the global model's test accuracy drops to approximately 10\% in the absence of any defense mechanism across all datasets, rendering the model ineffective. On simpler tasks such as MNIST and FMNIST, most robust aggregation schemes, including Median, Trimmed Mean, and Geometric Median, successfully mitigate the attack and achieve comparable test accuracy to our method. However, for more complex settings like ResNet-18 trained on the CIFAR-10 dataset, these methods struggle to counter the attack effectively. In contrast, only \textsc{Krum} and the proposed Bayesian Robust Aggregation method maintain high test accuracy, demonstrating the robustness of our method in handling complex tasks under adversarial conditions. For the \emph{Backdoor attack}, we observe results similar to the SignFlip attack. On simpler datasets (MNIST and FMNIST), most defense mechanisms mitigate the attack effectively, achieving minimal attack success rates (ASR). However, for more complex tasks like CIFAR-10, the adversary bypasses many defenses, leading to high ASR. Notably, \textsc{Krum} (with user-defined hyperparameter) and the proposed Bayesian Robust Aggregation scheme are the only methods capable of significantly suppressing the ASR while maintaining high test accuracy. 

\cref{tab:results-template1-signback-config2} presents the results for configuration 2, where the proportion of malicious clients increases to 40\%. As expected, the higher proportion of adversaries amplifies the challenges of maintaining model integrity. For the \emph{SignFlip attack}, many baseline aggregation methods experience a substantial drop in performance. For instance, Median and Trimmed Mean exhibit a 20\% decrease while Geometric Median exhibits a 10\% decline in test accuracy on CIFAR-10 compared to Configuration 1. In contrast, \textsc{Krum} and our method consistently outperform others, achieving test accuracies within 1\% of the non-adversarial baseline. For the \emph{Backdoor attack}, the increased proportion of malicious clients leads to higher ASR across most methods even for simpler tasks like MNIST and FMNIST. However, our Bayesian Robust Aggregation method maintains a low ASR (e.g., $<$10\% on CIFAR-10) while preserving high test accuracy, demonstrating its effectiveness even under extreme adversarial influence.

\cref{tab:results-template1-signback-config3} shows results for configuration 3, where 45\% of clients are malicious and exhibit dynamic behavior by attacking intermittently to evade detection. This setup represents a highly challenging adversarial environment. While most defenses fail to mitigate the attacks, our Bayesian Robust Aggregation method maintains low attack success rates and high test accuracy, demonstrating resilience against both static and dynamic adversarial strategies.

\subsection{Ablation Study}
\label{subsec:ablation}

Additionally, we run the ablation studies to, first, verify that our method does not degrade performance in non-adversarial settings compared with standard Federated Averaging; and secondly, test how the achieved accuracy varies based on the gradually increased percentage of compromised nodes in comparison with alternative schemes that rely on a hyperparameter. Thus, similar to configurations considered above, we use 20 clients in total and use $\varepsilon$ equal to 0\%, 10\%, 20\%, 30\%, and 40\% for each individual experiment under \emph{SignFlip} and \emph{Backdoor} attacks. The training is performed for a more challenging dataset, CIFAR-10, using the same model and optimization settings as in Configuration 1--3. However, for the alternative algorithms that require a hyperparameter, throughout the full series of experiments, we specify the conservative value: $\beta = 0.4$ for \textsc{TrimAvg} and $L = 12$ for \textsc{Krum}. \cref{fig:ablation-sign_back} shows the results in terms of test accuracy for \emph{SignFlip} and attack success rate for \emph{Backdoor}, averaged over the last 20 communication rounds (we do not define the attack success rate for $\varepsilon = 0$). Results for \emph{Random} and \emph{LabelFlip} attacks can be found in \cref{section:appendix-additional_results}.

We observe that in the absence of adversaries, our method achieves the same level of test accuracy as \textsc{FedAvg}, across all datasets, demonstrating that it maintains utility in benign environments while still providing robust defense against the attacks. Similarly to the results in \cref{tab:results-template1-fliprand-config3}, only \textsc{Krum} and our aggregation method successfully mitigate the Backdoor attack. Here, we note that the proposed approach does not rely on additional hyperparameters, which is a key contribution. Additionally, our algorithm attains high accuracy consistently throughout the entire range of the typical adversarial proportions.

\subsection{Limitations}
\label{subsec:limitations}
    It may be tempting to use the described variational inference based on the multivariate Gaussian distribution. Such modification could allow using different scale for each model weight individually and therefore better capture the deviations from an `honest' model update.  However, the shortcoming of the evidence lower bound involved in our derivations is that the KL-term $\mathrm{KL}(\weight, \varepsilon)$ has a fixed scale, while the multivariate Gaussian has the log-likelihood which scales linearly with the dimension of the model weigths. This leads to inconsistency in the setting of high dimension which is typical for Deep Learning models. Instead, we use the  Gaussian likelihood defined for scalar residuals $\mynorm{\bm{w}^t_k - \widehat{\bm{w}}}^2$ and thus avoid the dependency on the dimension. Interestingly, \textsc{Krum} is also based on the regular Euclidean distances, which is still sufficient to make it effective against the targeted attacks, such as \emph{Backdoor}, when the malicious updates are harder to detect.

\section{Conclusion}
We presented an adaptive robust aggregation approach for Federated Learning based on Bayesian inference. The mean update is defined as the maximum of the likelihood, marginalized over the probabilities of clients being honest. The approach only considers two assumptions -- the Huber model of contamination 
and the Gaussian likelihood to model residuals in the mean updates.
As a result, the proposed approach is simple in nature, and does not require specification of the number of compromised clients. We demonstrate the efficacy of our Bayesian robust aggregation approach on benchmark classification tasks in Federated Learning, where it consistently outperforms baseline defenses, particularly in complex tasks and high-adversary settings. The method demonstrates resilience against both static and dynamic adversarial strategies, maintaining high test accuracy and low attack success rates. We also examine the practicality of the method for real-world deployment by validation in the non-adversarial setting, where the method maintains comparable performance to the standard \textsc{FedAvg}.

\subsubsection*{Acknowledgements}
The computations were enabled by resources provided by the National Academic Infrastructure for Supercomputing in Sweden (NAISS), partially funded by the Swedish Research Council through grant agreement no. 2022-06725.

\clearpage
\bibliography{main}
\bibliographystyle{icml2025}

\clearpage
\appendix
\section{Detailed Experiment Setup}
\label{section:appendix-experiment-details}

\subsection{Datasets and Models}
\label{section:appendix-datasets-models}

To facilitate the replication of our results, we provide a detailed explanation of the experiment setup, including datasets, augmentations, models, and training parameters. Our code is publicly available at: (link removed for anonymity).

We conduct experiments on three classic image classification datasets: 
\begin{enumerate}
    \item \textbf{MNIST}~\cite{deng2012mnist}: A dataset of 60,000 grayscale images of handwritten digits, each of size $28\times28$ pixels, uniformly distributed across 10 classes. It is split into 50,000 training images and 10,000 testing images. We resize the images to $32\times32$ pixels using constant padding and normalize them using a mean $=0.1307$ and standard deviation $=0.1307$.
    \item \textbf{Fashion-MNIST}~\cite{xiao2017fmnist} dataset similarly contains grayscale images of size $28\times28$ pixels, representing 10 different classes of fashion items. It includes 60,000 training images and 10,000 testing images. Similar to MNIST, we resize the images to $32\times32$ pixels and normalize them using a mean $=0.5$ and standard deviation $=0.5$.
    \item The CIFAR-10~\cite{krizhevsky2009learning} dataset comprises 60,000 color images, each of size $32\times32$ pixels, uniformly divided into 10 classes, with 50,000 used for training and 10,000 for testing. We normalize the images using mean $= [0.4914, 0.4822, 0.4465]$ and standard deviation $= [0.2023, 0.1994, 0.2010]$ and apply random horizontal flip augmentation with a probability of $0.5$.
\end{enumerate}
For MNIST and Fashion-MNIST, we use the LENET-5 architecture, which consists of two convolutional layers followed by three fully connected layers. For CIFAR-10, we use a ResNet-18 model, a deep residual network with skip connections.

\subsection{Training Parameters}
\label{section:appendix-federated-setup}
\begin{table}[t]
    \vskip 0.1in
    \caption{Summary of parameters used for local client side training.}
    \label{tab:configs-t2-training}
    \vskip 0.1in
    \begin{center}
        \begin{small}
            \begin{tabular}{p{5cm}c}\toprule[1.5pt]
                \textbf{Parameter}          & \textbf{Value}        \\ \midrule
                 Learning rate ($\eta$)     &   $0.01$              \\ 
                 Learning rate Scheduler    &   No                  \\ 
                 Batch size ($B$)           &   $128$               \\ 
                 \# of Local Epochs ($E$)   &   $10$                \\ 
                 Optimizer                  &   SGD                 \\ 
                 Momentum                   &   $0.9$               \\ 
                 Weight decay (L2 Penalty)  &   $1\mathrm{e}{-4}$   \\ 
                 \bottomrule[1.5pt]
            \end{tabular}
        \end{small}
    \end{center}
    \vskip -0.1in
\end{table}

The training parameters for all experiments are summarized in \cref{tab:configs-t2-training}. We use Stochastic Gradient Descent (SGD) with L2 regularization, Nesterov momentum ($0.9$), and a weight decay of $1\mathrm{e}{-4}$. The initial learning rate is set to $0.01$ and is kept static throughout the training. We train the models for $10$ epochs with a batch size of $128$. 

\subsection{Hardware Details}
Each experiment is conducted on a single NVIDIA A40 GPU.

\subsection{Attack Details}
\label{section:appendix-atttack-details}
The specifications of each attack type are described below.
\subsubsection{Sign Flipping Attack}
The \emph{sign flipping attack} aims to disrupt the global model convergence by flipping the signs of model updates, causing the aggregated updates to diverge. Specifically, each compromised client modifies its gradient update $\mathbf{g}$ as follows:
\begin{equation}
    \mathbf{g}_{\text{malicious}} = -\gamma \times \mathbf{g}_{\text{honest}},
\end{equation}
where $\mathbf{g}_{\text{honest}}$ is the honest model update computed using local dataset, and $\gamma$ is the scaling factor used to amplify the impact of malicious update. For all experiments, we set $\gamma = 4.0$.

\subsubsection{Backdoor Attack}
\begin{figure}[t]
    \centering
    \begin{subfigure}[b]{0.33\columnwidth}
        \centering
        \includegraphics[width=0.75\columnwidth]{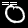}
        \caption{}
    \end{subfigure}%
    \begin{subfigure}[b]{0.33\columnwidth}
        \centering
        \includegraphics[width=0.75\columnwidth]{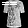}
        \caption{}
    \end{subfigure}%
    \begin{subfigure}[b]{0.33\columnwidth}
        \centering
        \includegraphics[width=0.75\columnwidth]{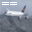}
        \caption{}
    \end{subfigure}
    \caption{Examples of backdoor-triggered samples from (a) MNIST, (b) Fashion-MNIST, and (c) CIFAR-10 datasets.}
    \label{fig:triggered-samples}
\end{figure}

For the \emph{backdoor attack}, our trigger design follows the specifications used by \cite{zhang2023a3fl}. The trigger consists of two parallel lines resembling a double equal sign (==), designed to be small yet effective. The details of the trigger design as as follows:
\begin{itemize}
    \item Horizontal gap between strokes: 1 pixel
    \item Vertical gap between strokes: 1 pixel
    \item Stroke width: 7 pixels
    \item Stroke height: 1 pixel
    \item Trigger position: Offset 2 pixels from the top-left corner of the image
\end{itemize}
A visual illustration of the triggered samples is provided in \cref{fig:triggered-samples}. For all datasets, we introduce the backdoor trigger to samples from class $0$ and relabel them as class $8$. The adversaries conducting the backdoor attack follow a two-step training process: first, they train the local model exclusively on poisoned data; then, they refine the model by further training it on a mix of benign and malicious data to restore high accuracy on benign tasks. Our observations indicate that using only the first step significantly degrades performance on benign tasks, while using only the second step results in a low attack success rate (ASR).

Additionally, for the CIFAR-10 dataset, we found that initiating the attack from the beginning of training prevents the model from converging. Therefore, following a similar strategy to Zhang \etal~\yrcite{zhang2023flip}, we allow the model to first converge for 200 communication rounds before launching the backdoor attack.

\subsubsection{Label Flipping Attack}
For the \emph{label flipping attack}, adversarial clients modify the labels of their training data to induce misclassification. The transformation follows a cyclic label shift:
\begin{equation}
    y_{\text{malicious}} \leftarrow y_{\text{honest}} + 1 \mod Y,
\end{equation}
where $Y$ is the total number of classes in the dataset. For example, in a 10-class setting (e.g., MNIST, CIFAR-10), label 0 is flipped to 1, label 1 to 2, and so on, with label 9 wrapping around to 0.

\subsubsection{Random Update Attack}
In the \emph{random update attack}, adversarial clients replace their model updates with random noise sampled from a Gaussian distribution. The noise variance is scaled according to the magnitude of the honest update. i.e.,
\begin{equation}
    \mathbf{g}_{\text{malicious}} \sim \mathcal{N}\left(0,\; \gamma \times \left| \mathbf{g}_{\text{honest}} \right|^2 \right),
\end{equation}
where $\gamma$ is scaling factor that ensures the noise is strong enough to disrupt the training while preserving a plausible update magnitude. For all experiments, we set $\gamma = 4.0$.

\begin{figure*}[t]
    \centering
    \begin{subfigure}[b]{0.45\linewidth}
        \centering
        \includegraphics[width=0.80\linewidth]{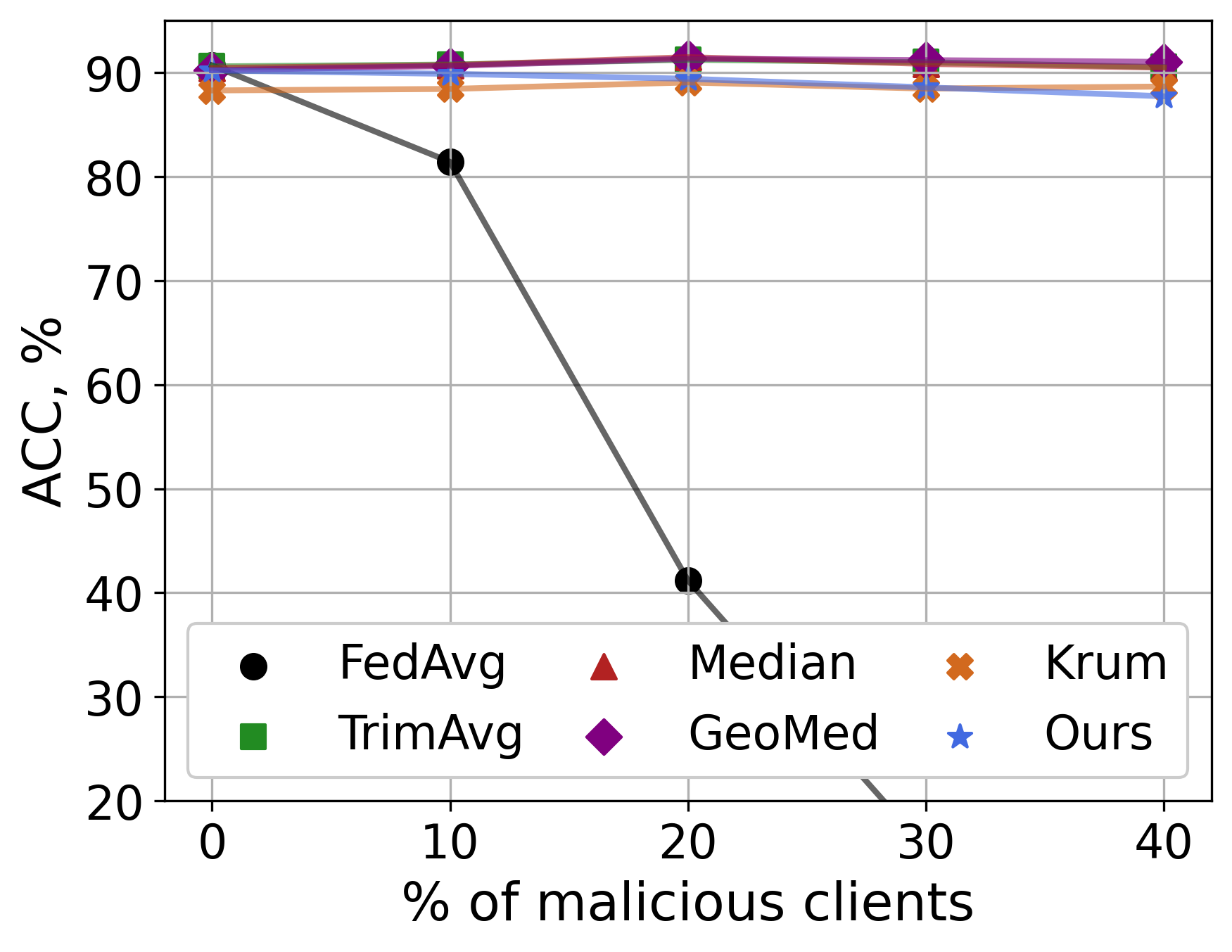}
        \caption{}
    \end{subfigure}%
    \begin{subfigure}[b]{0.45\linewidth}
        \centering
        \includegraphics[width=0.80\linewidth]{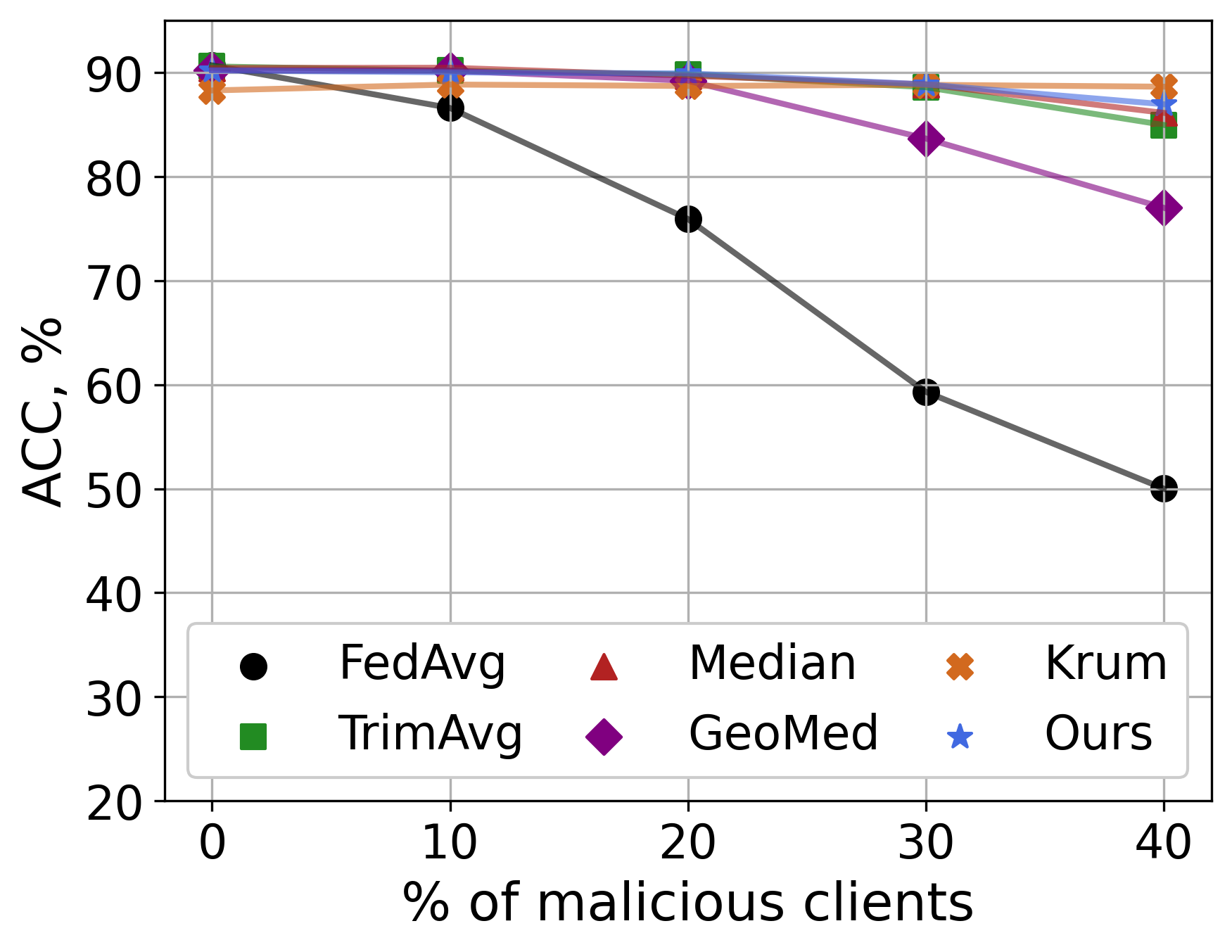}
        \caption{}
    \end{subfigure}\vfill
    \caption{Ablation study. Test accuracy (ACC) under (a) Random and (b) Label-flipping attacks, with fraction of malicious clients varying from 0 to 40\%.}
    \label{fig:ablation-flip_rand}
\end{figure*}

\begin{table}[h]
    \vskip 0.1in
    \caption{Results showing ACC of the global model under Label Flipping and Random attacks with $20\%$ malicious clients. \\
    $^\ast$We use $\beta = 0.2$ for \textsc{TrimAvg}, and $L = 16$ for \textsc{Krum}.}
    \label{tab:results-template1-fliprand-config1}
    \vskip 0.1in
    \begin{center}
        \begin{small}
        \resizebox{0.95\columnwidth}{!}{%
            \begin{tabular}{llgcgc}\toprule[1.5pt]
                & \multirow{3}{*}{\textbf{Baseline}} & \multicolumn{2}{c}{\textbf{LabelFlip}}  & \multicolumn{2}{c}{\textbf{Random}} \\
                \cmidrule(lr){3-4} \cmidrule(lr){5-6}
                 &  &  \multicolumn{1}{c}{$\alpha = 0.5$}  & $\alpha = 1.0$ & \multicolumn{1}{c}{$\alpha = 0.5$}  & $\alpha = 1.0$ \\ \midrule[1.125pt]
                \multirow{7}{*}{\rotatebox{90}{\textbf{MNIST}}} 
                & \textsc{FedAvg} & 0.82 & 0.85 & 0.82 & 0.83 \\
                ~ & \textsc{Median} & 0.99 & 0.99 & 0.99 & 0.99 \\
                ~ & \textsc{TrimAvg}* & 0.99 & 0.99 & 0.99 & 0.99 \\
                ~ & \textsc{GeoMed} & 0.99 & 0.99 & 0.99 & 0.99 \\
                ~ & \textsc{Krum}* & 0.99 & 0.99 & 0.99 & 0.99 \\ \cmidrule(lr){2-6}
                ~ & \textbf{Ours} & 0.99 & 0.99 & 0.99 & 0.99 \\ \midrule[1.125pt]

                \multirow{7}{*}{\rotatebox{90}{\textbf{FMNIST}}}
                & \textsc{FedAvg} & 0.65 & 0.70 & 0.76 & 0.76 \\
                ~ & \textsc{Median} & 0.89 & 0.89 & 0.87 & 0.88 \\
                ~ & \textsc{TrimAvg}* & 0.89 & 0.89 & 0.87 & 0.87 \\
                ~ & \textsc{GeoMed} & 0.89 & 0.89 & 0.88 & 0.87 \\
                ~ & \textsc{Krum}* & 0.89 & 0.89 & 0.89 & 0.89 \\ \cmidrule(lr){2-6}
                ~ & \textbf{Ours} & 0.89 & 0.89 & 0.89 & 0.89 \\ \midrule[1.125pt]
                 
                \multirow{7}{*}{\rotatebox{90}{\textbf{CIFAR-10}}}
                & \textsc{FedAvg} & 0.41 & 0.40 & 0.76 & 0.76 \\
                ~ & \textsc{Median} & 0.91 & 0.91 & 0.90 & 0.89 \\
                ~ & \textsc{TrimAvg}* & 0.91 & 0.91 & 0.89 & 0.89 \\
                ~ & \textsc{GeoMed} & 0.91 & 0.91 & 0.89 & 0.90 \\
                ~ & \textsc{Krum}* & 0.90 & 0.90 & 0.90 & 0.90 \\ \cmidrule(lr){2-6}
                ~ & \textbf{Ours} & 0.89 & 0.89 & 0.90 & 0.90 \\
                 \bottomrule[1.5pt]
            \end{tabular}%
        }
        \end{small}
    \end{center}
    \vskip -0.1in
\end{table}

\begin{table}[h]
    \vskip 0.1in
    \caption{Results showing ACC of the global model under Label Flipping and Random attacks with $40\%$ malicious clients. \\ $^\ast$We use $\beta = 0.4$ for \textsc{TrimAvg}, and $L = 12$ for \textsc{Krum}.}
    \label{tab:results-template1-fliprand-config2}
    \vskip 0.1in
    \begin{center}
        \begin{small}
        \resizebox{0.95\columnwidth}{!}{%
            \begin{tabular}{llgcgc}\toprule[1.5pt]
                & \multirow{3}{*}{\textbf{Baseline}} & \multicolumn{2}{c}{\textbf{LabelFlip}}  & \multicolumn{2}{c}{\textbf{Random}} \\
                \cmidrule(lr){3-4} \cmidrule(lr){5-6}
                 &  &  \multicolumn{1}{c}{$\alpha = 0.5$}  & $\alpha = 1.0$ & \multicolumn{1}{c}{$\alpha = 0.5$}  & $\alpha = 1.0$ \\ \midrule[1.125pt]
                \multirow{7}{*}{\rotatebox{90}{\textbf{MNIST}}} 
                & \textsc{FedAvg} & 0.42 & 0.36 & 0.51 & 0.52 \\
                ~ & \textsc{Median} & 0.99 & 0.99 & 0.98 & 0.98 \\
                ~ & \textsc{TrimAvg}* & 0.99 & 0.99 & 0.98 & 0.99 \\
                ~ & \textsc{GeoMed} & 0.99 & 0.99 & 0.99 & 0.99 \\
                ~ & \textsc{Krum}* & 0.99 & 0.99 & 0.99 & 0.99 \\ \cmidrule(lr){2-6}
                ~ & \textbf{Ours} & 0.99 & 0.99 & 0.99 & 0.99 \\ \midrule[1.125pt]

                \multirow{7}{*}{\rotatebox{90}{\textbf{FMNIST}}}
                & \textsc{FedAvg} & 0.29 & 0.31 & 0.48 & 0.48 \\
                ~ & \textsc{Median} & 0.89 & 0.89 & 0.82 & 0.83 \\
                ~ & \textsc{TrimAvg}* & 0.89 & 0.89 & 0.82 & 0.81 \\
                ~ & \textsc{GeoMed} & 0.89 & 0.89 & 0.61 & 0.62 \\
                ~ & \textsc{Krum}* & 0.89 & 0.89 & 0.89 & 0.89 \\ \cmidrule(lr){2-6}
                ~ & \textbf{Ours} & 0.89 & 0.89 & 0.83 & 0.85 \\ \midrule[1.125pt]
                 
                \multirow{7}{*}{\rotatebox{90}{\textbf{CIFAR-10}}}
                & \textsc{FedAvg} & 0.11 & 0.10 & 0.50 & 0.50 \\
                ~ & \textsc{Median} & 0.90 & 0.90 & 0.86 & 0.87 \\
                ~ & \textsc{TrimAvg}* & 0.91 & 0.91 & 0.86 & 0.86 \\
                ~ & \textsc{GeoMed} & 0.91 & 0.91 & 0.77 & 0.71 \\
                ~ & \textsc{Krum}* & 0.89 & 0.89 & 0.89 & 0.89 \\ \cmidrule(lr){2-6}
                ~ & \textbf{Ours} & 0.88 & 0.88 & 0.90 & 0.90 \\
                 \bottomrule[1.5pt]
            \end{tabular}%
        }
        \end{small}
    \end{center}
    \vskip -0.1in
\end{table}

\begin{table}[h]
    \vskip 0.1in
    \caption{Results showing ACC of the global model under Label Flipping and Random attacks with $\leq 45\%$ malicious clients. \\$^\ast$We use $\beta = 0.45$ for \textsc{TrimAvg}, and $L = 11$ for \textsc{Krum}.}
    \label{tab:results-template1-fliprand-config3}
    \vskip 0.1in
    \begin{center}
        \begin{small}
        \resizebox{0.95\columnwidth}{!}{%
            \begin{tabular}{llgcgc}\toprule[1.5pt]
                & \multirow{3}{*}{\textbf{Baseline}} & \multicolumn{2}{c}{\textbf{LabelFlip}}  & \multicolumn{2}{c}{\textbf{Random}} \\
                \cmidrule(lr){3-4} \cmidrule(lr){5-6}
                 &  &  \multicolumn{1}{c}{$\alpha = 0.5$}  & $\alpha = 1.0$ & \multicolumn{1}{c}{$\alpha = 0.5$}  & $\alpha = 1.0$ \\ \midrule[1.125pt]
                \multirow{7}{*}{\rotatebox{90}{\textbf{MNIST}}} 
                & \textsc{FedAvg} & 0.71 & 0.70 & 0.73 & 0.73 \\
                ~ & \textsc{Median} & 0.99 & 0.99 & 0.99 & 0.99 \\
                ~ & \textsc{TrimAvg}* & 0.99 & 0.99 & 0.99 & 0.99 \\
                ~ & \textsc{GeoMed} & 0.99 & 0.99 & 0.99 & 0.99 \\
                ~ & \textsc{Krum}* & 0.99 & 0.99 & 0.99 & 0.99 \\ \cmidrule(lr){2-6}
                ~ & \textbf{Ours} & 0.99 & 0.99 & 0.99 & 0.99 \\ \midrule[1.125pt]

                \multirow{7}{*}{\rotatebox{90}{\textbf{FMNIST}}}
                & \textsc{FedAvg} & 0.55 & 0.57 & 0.67 & 0.66 \\
                ~ & \textsc{Median} & 0.90 & 0.90 & 0.85 & 0.86 \\
                ~ & \textsc{TrimAvg}* & 0.90 & 0.89 & 0.85 & 0.85 \\
                ~ & \textsc{GeoMed} & 0.90 & 0.90 & 0.82 & 0.82 \\
                ~ & \textsc{Krum}* & 0.89 & 0.89 & 0.89 & 0.89 \\ \cmidrule(lr){2-6}
                ~ & \textbf{Ours} & 0.89 & 0.89 & 0.89 & 0.89 \\ \midrule[1.125pt]
                 
                \multirow{7}{*}{\rotatebox{90}{\textbf{CIFAR-10}}}
                & \textsc{FedAvg} & 0.25 & 0.26 & 0.67 & 0.67 \\
                ~ & \textsc{Median} & 0.92 & 0.91 & 0.89 & 0.89 \\
                ~ & \textsc{TrimAvg}* & 0.92 & 0.92 & 0.89 & 0.89 \\
                ~ & \textsc{GeoMed} & 0.91 & 0.92 & 0.87 & 0.86 \\
                ~ & \textsc{Krum}* & 0.89 & 0.89 & 0.90 & 0.89 \\ \cmidrule(lr){2-6}
                ~ & \textbf{Ours} & 0.90 & 0.90 & 0.90 & 0.90 \\
                 \bottomrule[1.5pt]
            \end{tabular}%
        }
        \end{small}
    \end{center}
    \vskip -0.1in
\end{table}

\section{Additional Results}
\label{section:appendix-additional_results}

Tables \ref{tab:results-template1-fliprand-config1}, \ref{tab:results-template1-fliprand-config2}, and \ref{tab:results-template1-fliprand-config3} present the testing accuracy (ACC) of the global model under Label Flipping and Random Update attacks with varying percentages of malicious clients (20\%, 40\%, and up to 45\%). Across all configurations, the baseline methods and our proposed approach show strong performance, with only minor fluctuations observed in the results.

For Random Update attack, methods like \textsc{FedAvg}, \textsc{Median}, \textsc{TrimAvg}, and \textsc{Krum} remain robust, achieving near-perfect accuracy in most cases, with slight degradation observed at higher levels of malicious clients. Similarly, in the case of the Label Flipping attack, all methods, including ours, exhibit stable performance. This suggests that while these attacks can still have some impact, they do not drastically affect the models' overall accuracy.

The consistency of the results across different datasets (MNIST, FMNIST, CIFAR-10) and attack configurations further supports the conclusion that these attack types do not challenge the robustness of the methods in the same way as more sophisticated attacks, such as Sign Flipping or Backdoor attack. Given the minimal impact of these attacks on model accuracy, these results are primarily reported for completeness rather than to highlight significant differences in performance.

\newpage
\section{Formal Analysis}
\label{section:appendix-analysis}

In the main text we state (\cref{proposition}) that the aggregation rule that returns a centroid $\overline{\bm{w}}_S$ of the points in the subset defined by
\begin{align}
    \min_{\substack{S \subseteq \{1,...,K\} \\ \vert S \vert = K-M}} \sum_{k \in S} \Vert \bm{w}_k - \overline{\bm{w}}_S \Vert_2^2,
    \label{eq:set-form-app}
\end{align}
is $(M, \kappa)-$robust. 

Below we provide the proof of \cref{proposition}; it follows the argument similar to the proof for Krum in \cite{allouah2023fixing}. 
\begin{proof}
Consider any set $S \subseteq \{1, ..., K\}$, s. t. $\vert S \vert = K-M$.
Using Jensen's inequality, we have $\forall \, k \in B$,
\begin{align}
\Vert \overline{\bm{w}}_S - \overline{\bm{w}}_B \Vert_2^2 \leq 2 \Vert \bm{w}_k - \overline{\bm{w}}_S \Vert^2 + 2 \Vert \bm{w}_k - \overline{\bm{w}}_B \Vert_2^2.
\end{align}
Therefore,
\begin{align}
    &\sum_{k \in S} \Vert \bm{w}_k - \overline{\bm{w}}_S \Vert^2 \geq \sum_{k \in S \cap B} \Vert \bm{w}_k - \overline{\bm{w}}_S \Vert_2^2 \\
    &\geq \frac{\vert S \cap B \vert }{2} \Vert \overline{\bm{w}}_S - \overline{\bm{w}}_B \Vert^2 - \sum_{k \in S \cap B}\Vert \bm{w}_k - \overline{\bm{w}}_B \Vert_2^2.
    \nonumber
\end{align}
Rearranging the terms and using 
\begin{align}
    \vert S \cap B \vert = \vert S \vert + \vert B \vert - \vert S \cup B \vert \geq K - 2M,
\end{align}
we obtain
\begin{align}
&\Vert \overline{\bm{w}}_S - \overline{\bm{w}}_B \Vert^2 \leq \\
&\leq \dfrac{2}{K-2M} \left( \sum_{k \in S} \Vert \bm{w}_k - \overline{\bm{w}}_S \Vert^2 + \sum_{k \in S \cap B} \Vert \bm{w}_k - \overline{\bm{w}}_B \Vert^2 \right) 
\nonumber
\\
&\leq\dfrac{2}{K-2M} \left( \sum_{k \in S} \Vert \bm{w}_k - \overline{\bm{w}}_S \Vert^2 + \sum_{k \in B} \Vert \bm{w}_k - \overline{\bm{w}}_B \Vert^2 \right)
\nonumber
\end{align}
This is true for any subsets $S$ and $B$ of cardinality $K-M$. By definition, if $S$ is a solution of \cref{eq:set-form-app}, we have $\forall \, S^\prime \subseteq \{1,..., K\}, \, \vert S^\prime \vert = K-M$,
\begin{align}
\sum_{k \in S} \Vert \bm{w}_k - \overline{\bm{w}}_{S} \Vert^2 \leq \sum_{k \in S^\prime} \Vert \bm{w}_k - \overline{\bm{w}}_{S^\prime} \Vert_2^2
\end{align}
including $S^\prime = B$. Thus, given a solution $S$ of \cref{eq:set-form-app}, we obtain the desired result for the aggregated vector $\overline{\bm{w}}_{S}$:
\begin{align}
\Vert \overline{\bm{w}}_{S} - \overline{\bm{w}}_B \Vert^2 \leq \dfrac{4 (K - M)}{K-2M} \dfrac{1}{\vert B \vert}\sum_{k \in B} \Vert \bm{w}_k - \overline{\bm{w}}_B \Vert_2^2
\end{align}
Therefore, $\kappa = 4\dfrac{K-M}{K-2M} = 4\left(1 + \dfrac{M}{K-2M}\right)$.
\end{proof}

\end{document}